# A data science axiology: the nature, value, and risks of data science

Michael L. Brodie mlbrodie@seas.harvard.edu
Data Systems Laboratory, School of Engineering and Applied Sciences
Harvard University, Cambridge, MA USA

=============DRAFT July 18, 2023=====================

Data science is not a science. It is a research paradigm. Due its power, scope, and scale, it will surpass science – our most powerful research paradigm – in enabling knowledge discovery that is changing our world[10]. This paper explores and evaluates its remarkable, definitive features.

Modern data science is in its infancy. Emerging slowly since 1962 and rapidly since 2000, data science is a fundamentally new field of inquiry, one of the most active, powerful, and rapidly evolving innovations of the 21st century. Due to its value, power, and scope of applicability, it is emerging in over 40 disciplines, hundreds of research areas, and tens of thousands of applications. Yet we are just beginning to understand and define it. $10^6$ data science publications contain myriad definitions of data science and data science problem solving. Due to its infancy, many definitions are independent, application-specific, mutually incomplete, redundant, or inconsistent, hence so is data science as a field of inquiry. This research addresses this data science multiple definitions challenge by proposing the development of coherent, unified definition of data science based on a data science reference framework[46]-[53] by means of a data science journal[54] for the data science community to achieve such a definition.

This paper presents an axiology of data science, its purpose, nature, importance, risks, and value for problem solving based on its unique ability to computationally analyze data to discover insights into motivating domain problems where the scope, scale, and complexity of problems and solutions can be beyond human capacity to understand and discoveries not otherwise possible. A comprehensive data science axiology is a theory of value that defines the nature, value, and uses of data science. As data science is in its infancy, its axiology can only be speculated. Such an axiology can aid in understanding and defining data science and recognizing potential benefits, risks, and open research challenges. We present the history and nature of data science and offer candidate definitions of essential data science concepts required to discuss its axiology. Within a decade, this remarkable new research paradigm will be seen as a milestone in human knowledge discovery by non-human means.

The Age of AI is changing our world practically and profoundly. Visionaries[11][15][26][34][35] herald data science as a *Promethean Moment*[10] that changes *everything*, requiring reimagining and restructuring our world. Unlike past moments based on single inventions, e.g., the printing press, this moment is based on a meta-technology[1] augmented by data of the whole world in what Tom Friedman calls *The Age of Acceleration, Amplification,* and *Democratization[10]* with impacts far beyond knowledge discovery.

1. **Introduction**

*1.1. Essential data science concepts*

Data science (the *data science research paradigm*) is defined by the philosophy of data science and the *data science reference framework* (axiology, ontology, epistemology, methodology, methods, technology) based on classical concepts[46]. To contribute to an initial assessment and definition of data science, this paper proposes an initial axiology of data science.

The *philosophy of data science* is the worldview that provides the philosophical underpinnings (i.e., learning from data) of data science research for knowledge discovery with which to understand, reason about, discover, articulate, and validate[2] *insights into the true nature of the ultimate questions about a phenomenon by computational analysis of a dataset that represents features of interest of some subset of the entire population of the phenomenon.* Data science results are *probabilistic, correlational, possibly fragile or specific to the analysis method or dataset, cannot be proven complete or correct, and lack explanations and interpretations for the motivating domain problem*[46].

---

[1] A meta technology is used to produce new technology and knowledge hence can be applicable to most human endeavors.

The purpose of *data science* is knowledge discovery conducted by applying the *data science problem solving paradigm* to data science problems. Hence, the axiology of the data science concerns data science problems, the data science problem solving paradigm, its methodology (the data science method, the data science workflow, governing principles), and methods (computational analytical methods, operands, solutions, results, governing principles) for knowledge discovery.

The widespread adoption of data science and its remarkable successes in contributing to solving significant, practical problems, e.g., protein folding[17], have led to much speculation by experts in research, application domains, business, and governments and to a deluge of speculation on data science with the theme that it will transform the world[4][10][11][19][20][21][34][35]. Expert assessments are readily available and will continue to emerge for decades.

### 1.2. Data science is not a science

As science is well understood, it is used to contrast with data science. A significant challenge of data science is the false understanding that it is a scientific pursuit as suggested in[8][9][36][41][42]. Data science benefits cannot be achieved by treating data science problem solving and results as scientific; rather it poses risks. Science and data science are peer research paradigms with mutually conflicting philosophies and results. In contrast with the philosophy of data science and its results (§1.1), the *philosophy of science* is the worldview that provides the philosophical underpinnings (i.e., objective, quantitative reasoning) of empirical research for knowledge discovery with which to understand, reason about, discover, articulate, and validate *the true nature of the ultimate questions about natural, observable phenomena as new knowledge about those phenomena.* Scientific results are *definitive, conclusive, casual, robust, universal knowledge of the phenomena with verified explanations and validated interpretations that are not provably complete*. Like all research paradigms, science and data science are complementary. Scientific results are used in data science and *vice versa*, just as science is used in archeology.

The scope of science is observable, measurable phenomena versus the scope of data science that is any phenomena with adequate data to support computational analysis. Data science can address problems of scope, scale, and complexity beyond human understanding, hence its unfathomed scope, scale, and power will surpass science as a knowledge discovery paradigm. While science surpasses data science in *knowing* due to the inscrutability of AI-based data science, data science surpasses science in *learning*.

### 2. An axiology of data science

The data science axiology is organized around the nature, benefits, and challenges of its definitive features. Most benefits and challenges are due to multiple features. To avoid repetition, each is described once. Some are intrinsic – concern data science *per se,* most are extrinsic – concern means for solving domain problems.

### 2.1. Data science is inherently risky

Due to the nature of data science *application risks* arise for every data science problem, solution, and result. It is difficult to evaluate or bound the risks of data science applications, requiring special considerations for life-, medical-, societal-, business-, environment-critical applications.

---

[2] Validation in science is inherent in the scientific method. Validation is impossible in data science, posing explanation and interpretation challenges.

Despite potential risks, data science has been widely adopted and applied due to its remarkable ability to provide insights that lead to pragmatic solutions for a wide range of domain problems. Such solutions and their results are successful only because plausible explanations of the solutions and acceptable interpretations of the results have been established, typically outside data science, to demonstrate that applying the solution and results for the motivating domain problem in practice is safe with risks within acceptable bounds. There are many techniques for creating such explanations and interpretations, none of which can prove the absence of risks let alone estimate risks in specific cases. Consequently, data science results are not appropriate for some problems, e.g., nuclear power plant failure predictions since no dataset contains enough information, and life-critical medical problems in which no risk is acceptable. Nevertheless, data science can be and has been used very successfully to provide insights into the motivating domain problems and results.

Due to the inherent risks of data science, practitioners should consider the Hippocratic oath –*First, cause no harm*– and apply only demonstrable, standard AI solutions[46] with acceptable bounds. Given the above properties inherent in data science reasoning and results, consider the challenge to ensure that military data science applications that could kill meet United Nations 2019 guidance that all decisions to take human life must involve human judgment and a [US Department of Defense 2020](#) requirement that its AI systems be responsible, equitable, traceable, reliable, and governable. As data science is devoid of such properties they must be evaluated and managed by humans.

### 2.2. *Data science is in its infancy*

Unlike science that took 400 years to formalize, data science may take only decades to mature and be better understood.

The chief benefit of the data science research paradigm is that it is a fundamentally new field of inquiry independent of other fields of inquiry that can lead to solving problems that cannot be solved otherwise, such as protein folding[17], an achievement that Eric Schmidt says has "changed science forever"[19] and, suggesting its profound potential "will change the very nature of scientific research"[35]. Despite being in its infancy, much has been learned and achieved in its 60-year development. In the Age of AI, scientists must pose the right questions, express and solve the corresponding AI problems, and develop explanations and interpretations that ensure safe, ethical use.

Significant challenges due to its infancy are reflected in those in the infancy of science in the 17th and 18th centuries. In the 1830s, Comte observed[5] that science's philosophical definition provided inadequate guidance for its conduct in scientific disciplines. It too faced the multiple definitions challenge thus was inadequately understood and defined[46] as is data science today. How do you understand and define a new field of inquiry? What are the underlying concepts and theories? Its many domain- or application-specific successes (techniques, methods, solutions) await generalizing up to data science followed by specialization down to relevant data science disciplines. Massive efforts in data science research – one of the most extensive in the world – are addressing these issues.

### 2.3. *Data science is exploration via data*

As envisaged by John Tukey, data science's originator, computational analysis of datasets can discover "what lies behind the data"[41][42] in scope, scale, and complexity that Tukey could not have imagined. The wealth of data science methods and datasets on critical features of most phenomena enable its chief purpose and benefits – extensive exploration of data denoting phenomena for insights into the truth of the ultimate questions about those phenomena[46].

### 2.4. *Data science is indirect*

The scientific research paradigm provides means for discovering, explaining and validating, and interpreting and verifying new knowledge about phenomena, i.e., definitive answers to scientific questions. Scientific analyses (experiments) are direct in three ways. They directly examine the phenomenon being analyzed; explanations of an analysis are inherent in the experimental design; and results interpretation are inherent in the hypotheses that experiments are designed to evaluate – a cause produces the hypothesized effect or it does not. The design and execution are verified by humans, I.e., scientist conducting the experiment and reviewing the results.

In contrast, data science analyses are indirect in the same three ways. First, it indirectly analyzes a phenomenon by analyzing a dataset believed to represent the features of the phenomenon critical to the analysis of a subset of the population of the phenomenon. Ideally the dataset is representative of the full population of the phenomenon but that is unknown and improbable. Second, it does not provide an answer to the motivating domain problem. Data science results are used as insights into the data science solution, i.e., trained data science model, that must be translated to insights into the motivating domain problem, solution, and result. Third, data science results are at best probabilistic evidence of patterns of features of the phenomenon. While data science results are evidence-based, i.e., patterns discovered directly in the observational dataset, they are of uncertain meaning and value. As data science results are computationally derived, they are correlational and not necessarily casual. Data science method algorithms can fail, e.g., image recognition models can fail due to adversarial data and possibly with non-adversarial data; therefore, results can be fragile, i.e., neither robust nor reliable. Worse, data science can hallucinate – produce results that appear rational and relevant but are wrong or fictitious. Due to their inscrutability, results cannot be proven complete or correct. To be applied in practice, these limitations must be assessed and quantified, e.g., by demonstrable, standard data science solutions[46] using means outside data science, e.g., empirical.

The value of data science to gain practical insights into domain problems has been demonstrated by applications in hundreds of disciplines and tens of thousands of applications, most that could not have been solved otherwise. The successes have raised data science to world-wide recognition. The potential scope, scale, and complexity of such solutions exceed those of all research paradigms, especially science.

Following the data science method [46] solves a motivating domain problem indirectly. First, the motivating domain problem must be translated into one or more data science problems. Second, the data science problems must be solved to meet specific requirements, e.g., accuracy. A computational method must be selected, trained, and tuned. A relevant dataset must be discovered, acquired, and prepared. The trained data science model is applied to the prepared dataset to produce a data science result. Finally, the data science problems, solutions, and results are used by humans to develop 1) an acceptable explanation that the analysis adequately implemented the intended data science and corresponding domain analyses, and 2) an acceptable interpretation of the data science results in terms of the motivating domain problem. Despite many such techniques, e.g., foundation[2] and world models, data science problem solving is a challenging art involving data science, domain, and problem solving knowledge and expertise.

### 2.5. *Unfathomed scope, scale, complexity, and power*

Data science can address domain problems at unfathomed scope, scale, and complexity with unfathomed benefits and risks, if the domain problems can be translated to corresponding data science problems. *Unfathomed* means that they can be of enormous scale with unknown limits due to data science's inscrutability and infancy.

*Unfathomed scope of applicability*: Data science is a general purpose, meta technology. It can be applied to any phenomenon for which there is adequate data, the domain problem can be transformed to a data science problem, and the data science solution and result can be used to gain insights into the motivating domain solution and results.

*Unfathomed power:* The power of data analyses depends on the combination of the computational method, the training and tuning of the method to produce a data science model with which to analyze datasets that are available and adequately represent the features and the phenomenon being analyzed, and methods for preparing datasets for the intended analysis. The enormous and rapidly growing number of extraordinarily successful applications in countless problem domains and widespread adoption demonstrates the unfathomed power of data science that exceeds all research paradigms.

*Unfathomed scale and complexity*: Data science can address problems, solutions, and results of unfathomed scale and complexity. Google's Switch Transformer exceeds 1.6T parameters. China's Wu Dao 2.0 transformer model has 1.75T parameters. The BaGuaLu model has 174T. AlphaGo has beaten the world master of Go, a game with $10^{170}$ board configurations.

Human concepts and reasoning are at small scope, scale, and complexity. To manage scope, scale, complexity in understanding, reasoning, and problem solving, concepts and problems are decomposed into manageable sizes. Reasoning in science is limited within Cartesian scientific disciplines. Scientific analyses are limited to N parameters, variables, or conditions as such experiments require evaluating $2^N$ cases. Most PhDs involve N < 4 (16 cases). Most experiments involve less than N=10 (1024 cases).

While this argument is based on the scientific method, it applies equally to all human reasoning, e.g., mathematics, and all research paradigms to bound the complexity of the problems addressed. Human knowledge is divided into Cartesian disciplines with many levels of sub-disciplines. A simple test of the human capacity for reasoning in complexity is to attempt to reason in four or five dimensions. Another is to consider planet earth and our sun as two of 200B celestial bodies in the Milky Way, then conceive of the Milky Way as one of 200B galaxies in the observable universe. Those scales, comparable to GPT-4's parameters, are inconceivable by humans. The terms *reasoning*, *considering*, *learning*, and *conceiving* are used to refer to computations despite our not understanding what the computations do.

Benefits of the unfathomed scope, scale, complexity, and power of data science is that it provides fundamentally new means for reasoning and analysis of problems vastly beyond the capacity of humans and science, hence not otherwise possible. Data science is mankind's next step in expanding knowledge discovery and knowledge. This profound speculation is made by many leading experts, e.g., "When it comes to very powerful technologies—and obviously AI is going to be one of the most powerful ever—we need to be careful." Demis Hassabis, DeepMind CEO [29].

### 2.6. *Unfathomed risks*

The remarkable benefits of the unfathomed scope, scale, complexity, and power of data science are juxtaposed with unfathomed risks. Due to the inscrutability of AI-based data science methods we lack a theory with which to understand and manage properties of data science solutions and results, e.g., prove properties such as accuracy, robustness, and reliability. This makes the application of such results to the motivating domain problem inherently risky. If data science is inherently inscrutable, yet to be determined, no such theory is possible. Lacking a theory, evaluating such properties is often done empirically, e.g., to find optimal data representations, and training and tuning methods.

The extent of such risks can be estimated by the nature of the application, e.g., recommending a movie versus a medical treatment. The *data science application risk* problem might assume that the data science application is intended to create positive outcomes – to be used for good with risks arising from failing to solve the domain problem with a positive outcome. As a *dual use* technology, data science's unfathomed power for positive benefits can be used to harm and destroy. Education and guardrails are required. "Not everybody is thinking about those things. It's like experimentalists, many of whom don't realize they're holding dangerous material." Demis Hassabis, CEO of DeepMind[29]. Consider shocking examples from world class experts.

The extreme risk of nuclear war due to the use of data science is identified by America's preeminent living statesman Henry Kissinger and Google's past-CEO, Eric Schmidt in[11] and elaborated in The Age of AI: And Our Human Future [19].

The potential of data science to destroy is illustrated by Collaborations Pharmaceuticals, Inc.'s MegaSyn system that uses deep reinforced learning to learn new classes of low-toxicity molecules that humans would not have otherwise discovered, as candidates for new drugs[43]. The Swiss Federal Institute for NBC (nuclear, biological, chemical) Protection enquired of the potential of MegaSyn to produce toxic molecules. By reversing the reinforcement reward, six hours of computing produced 40,000 toxic molecules that included the known most toxic molecules and more that humans had not yet discovered. Collaborations Pharmaceuticals chose to never disclose the solution.

Michael Jordan, a leading statistician and data scientist, observed[15][16] that the pervasive use of machine learning (ML) in systems could lead to unanticipated consequences and risks. ML-based systems are in widespread use in medicine and healthcare. While each application may be useful and innocuous on its own, what are their systemic or collective effects? Jordan envisioned "A planetary-scale inference-and-decision-making medical system will analyze and make recommendations on many medical conditions of patients for a very large number of patients. It will have all the relevant data and knowledge from medical research and practice. It will operate at scale in many aspects: knowledge, data, patients, doctors, medical conditions, functionality, etc. Such systems would provide the best solutions available in all domains, applying the right model for solving a problem, ensuring the correct data is used, etc.".

The Age of AI[19] suggests that data science is potentially more dangerous than previous dual use technologies due to three characteristics never previously held by a single technology: 1) dual use, i.e., the power to create and destroy, 2) potential for substantial destruction, and 3) easily adopted and deployed with limited resources. The mere public knowledge of Collaborations Pharmaceuticals, Inc.'s MegaSyn may be enough to be readily replicated with less ethical results.

### *2.7. Ambiguity and uncertainty*

For centuries, educated human reasoning – philosophy, mathematics, science – has sought outcomes that are certain, i.e., knowledge that, like scientific results, is *definitive, conclusive, casual, robust*, and *universal*, *with verified explanations and validated interpretations[46]*. Such reasoning and knowledge suggests a human desire for certainty. Despite life and our world being inherently uncertain, probabilistic means for reasoning with uncertainty are not commonly taught nor applied. Those with a lifelong experience of logical, scientific reasoning may be unfamiliar with uncertainty and uncomfortable with ambiguous, probabilistic reasoning, let alone at scales beyond human comprehension.

In contrast, data science reasoning is exclusively and inherently uncertain, ambiguous, and at scale. Data science results are *probabilistic, correlational, possibly fragile or specific to the analysis method or dataset, cannot be proven complete or correct, and lack explanations and interpretations[46]*. Uncertainty is inherent in AI-based data science methods, i.e., not provably accurate, reliable, nor robust. Datasets cannot be fully representative of the features of the phenomenon being analyzed nor of the entire population of the phenomenon. Hence, the pattern being sought may not be in the dataset or may not be present significantly to reflect its value, whatever *significance* means to the data science method and to the human interpreting the results.

The value of ambiguous and uncertain data science results can be substantial. First, the world is inherently uncertain. Data science contributes to framing and analyzing inherent uncertainty, i.e., more realistic than science that concludes by scientific induction that scientific knowledge is universally true, which even the fathers of science deemed implausible[14]. Probabilistic answers may well be more plausible and real than discrete scientific knowledge. Second, data science results depend on analyzed, incomplete datasets, so that retraining or inference over new datasets can produce correspondingly different outcomes, as is intuitive in a changing world. Third, data science results lead human problem solvers to insights that with work and non-data science means can lead to solutions to the motivating domain problem that could not have been achieved otherwise, hence exceed human problem solving performance. AlphaFold[17] is a data science solution for protein folding (a 50-year grand challenge) to discover new drugs and new classes of drugs. AlphaGo beat the world champion Go player. The data science research paradigm is taking mankind into new, previously inaccessible realms of problems and solutions with unfathomed consequences – into our Promethean AI moment.

The greatest challenge due to data science being ambiguous and uncertain is our discomfort with and lack of understanding, education, and practice with reasoning in uncertainty. Many people, including Einstein, by his own admission, are not comfortable with, educated or practiced in reasoning in uncertainty with uncertain, probabilistic, ambiguous outcomes. This can inhibit understanding data science – its purpose and analyses – and interpreting data science results. Despite data science and science being antithetical, it was named a science long ago[46] and by leading *data scientists*[8][9][36][41][42]. The implied certainty is unwarranted but the name stuck.

### 2.8. *Data science accelerates discovery*

In the early 21$^{st}$ century, the availability of massive, powerful, fast computing resources, and massive amounts of data led to many substantial successes in applying data science that in turn launched data science as the most active field of inquiry in research and practice. Data science's computational nature combined with massive computational resources enabled data science to accelerate discovery. A dramatic example was Moderna's spikevax COVID-19 vaccine that was designed in 48 hours using an existing ML-based mRNA vaccine development platform. Prior to the ML-based solution, drug design based on protein folding took a PhD-level team six to ten years suggesting Kurzweil's Singularity[3] may arrive early.

The chief benefit of accelerating discovery is that together with other benefits, it qualitatively changes scientific discovery as observed by Eric Schmidt[19][35] and DeepMind's mission statement "*We're solving intelligence to advance science and benefit humanity. At the heart of this mission is our commitment to act as responsible pioneers in the field of AI, in service of society's needs and expectations*". DeepMind has delivered on this mission statement spectacularly in less than 5 years with AlphaGo (Go), AlphaZero (chess, shogi), AlphaStar (multi-person strategy games), AlphaFold (protein folding), AlphaCode (coding engine). Those capabilities transform any discipline that meets requirements for applying data science. Accelerating discovery does not overcome any data science challenges, e.g., the application risk. Its scope accentuates them.

### 2.9. *Data science is largely inscrutable*

The purpose of a research paradigm is to provide means with which to reason (discover, analyze), articulate (explain, interpret), confirm (verify, validate), authorize, and curate (unify, organize) knowledge of phenomena in disciplines within the paradigm[46]. Data science exceeds all research paradigms in the scope, scale, and complexity of problems, solutions, and results addressed, yet it is inscrutable – we do not know how it works. Inscrutability results in data science providing no guidance on 1) *solution explanations* that explain how the problem was solved, required to verify that the intended analysis was conducted successfully; and 2) *analytical result interpretation* required to solve the motivating domain problem.

---

[3]At SWSX 2017, Kurzweil predicted that by 2045, the pace of technological change will be so rapid, its impact so deep, that human life will be irreversibly transformed. He envisaged discovery and progress "exploding with unexpected fury."

To be applied in practice, solutions to motivating domain problems require confirmation that explanations and interpretations are valid, verified, safe, and trustworthy. This is impossible in data science. The need for confirmation is determined by the risks posed by applying the explanation and interpretation in practice. There are many techniques to develop and confirm, to some degree, explanations and interpretations. The objective is to establish *demonstrable, standard data science solutions* within the research paradigm's problem solving paradigm that uses standard models to frame the domain and analysis problems. Guidance for doing so can follow the elegance and power of the scientific method that builds explanations and interpretations into the analysis method, i.e., validated empirical designs are explanations and verified conduct of a validated experiment is the interpretation.[46]

A consequence of the inscrutability of data science is that results must be interpreted as insights into the motivating problem that may be amenable to a range of solutions rather than a specific solution that is not broadly applicable or a broadly applicable solution that is not point-wise applicable. Such problem solving leads to attempts to understand the problem and a range of solutions which in turn motivates the development of theories and techniques to address the need to develop and confirm explanations and interpretations of data science solutions and results, and to better understand, approximate, evaluate, and manage accuracy, correctness, robustness, transparency, interpretability, explainability, and 90 more properties of data analyses[50][53].

Diminishing inscrutability is one of the greatest challenges in data science. Inscrutability may be inherent in data science, never to be understood. However, twenty years of research has provided progress on overcoming previously unknown aspects, e.g., demonstrable, standard data science solutions, foundation models, and better understanding data science models and data representations through empirical studies. A few decades of research may contribute to an emerging theory of data science. Science is full of overcoming inscrutable challenges, e.g., how elementary particles attain weight (Higgs boson), how atoms work (quantum mechanics), how human life is sustained (genes, DNA, central dogma of microbiology), and what holds the universe together (dark matter and dark energy).

### 2.10. Ethical use of data science

Data science's unfathomed scope of applicability means that its scope of ethical issues exceeds those of all other research paradigms. Two of many such issues are as follows.

The significant social challenge to ensure that data science applications are ethical came to worldwide prominence when Timnit Gebru, an AI ethics researcher, was summarily fired by Google for publishing ethical challenges of large language models[1]. As Google declared itself to be AI first[4], Gebru's concern was that Google would deploy AI without adequate understanding or managing the ethical challenges. As data science is inscrutable, its ethical deployment and use cannot be guaranteed without considerable effort almost entirely outside data science. A standard response to such issues is an AI ethics framework of which there are over 125 in Europe alone, including the Abrahamic Commitment by Catholics, Muslims, and Jews. AI ethics frameworks do not solve ethical problems, they define ethical objectives. There are currently no technical solutions. Human oversight, the only genuine solution, does not scale. Most large institutions and governments have declared AI as a major strategy, contributing to the AI arms race and posing the AI challenge at scale.

Another class of ethical challenges of data science are economic and political. The knowledge, resources, and cost to develop data science solutions for large-scale, complex domain problems, e.g., foundation models[2], limits their development and control to organizations with adequate resources, such as Big Tech, e.g., Alphabet (Google), Amazon, Microsoft, Apple, Meta (Facebook), and some governments. Such concentrations have historically led to economic and political challenges.

### 2.11. Data science can improve human decision making

Government, corporate, and personal decision making is typically based on intuition and expert opinion ideally supported by compelling evidence that leads to and supports the decision. Such decision making is often caricatured as deficient due to methods like grey-beards making critical decisions based on inadequate, unrepresentative evidence due to urgency, intuition (assuming adequate intelligence), experience (assuming adequate memory), instinct (shooting from the hip, hunches), political expediency (perhaps counter to or despite the facts), overwhelming heterogenous data, and worse. Research supports the caricature. Collective decision making, e.g., elections, surveys, social media, suffer similar challenges. Automation and digitization has attempted to improve decision making with tools and information systems. The continuous emergence of new such tools suggests their inadequacies – none of which had data science capabilities until 2020, when, by magic, a deluge of software products, including all web search engines and Microsoft's entire software suite, claimed such capabilities.

A great strength of data science is that, designed, conducted, and interpreted appropriately, data science results are data-driven, also called fact-based and evidence-based, thus providing factual evidence for human decision making. Data science can aid human decision making by expressing the decisions, i.e., domain problems, as data science problems that when applied to the relevant datasets, can provide insights by discovering evidence for or against the decision. Data science, applied diligently, directly addresses the inherent uncertainty in human decision making by encouraging – exploration of alternatives in formulating and discovering alternative decisions; analysis of various partitions of the population for sensitivity analysis – at an unfathomed scope, scale, complexity, power, and speed beyond all other knowledge discovery methods, devoid of, i.e., not limited by, human concepts and reasoning. These benefits have been realized in thousands of applications, e.g., accelerating discovery the Higgs boson in the largest physics dataset ever created; accelerating discovery of COVID-19 vaccines saving millions of lives; discovering under-served cohorts (pregnant women, racial groups) by existing COVID-19 vaccines leading to new vaccines in record time, saving further millions.

A 2011 political movement for data-driven decision making in the US government led to the [House Committee on Oversight and Government Reform (2015)](#) that states: *Without evidence, the federal government is an ineffective fiduciary on behalf of the taxpayer. Unfortunately, in many instances, federal decision-makers do not have access to the data necessary to best inform decisions. In such instances, agencies are unable to show the benefits or impacts of the programs they administer and cannot determine what, if any, unintended consequences are created by programs, or whether programs can be improved*. US public law H.R.4174, Foundations for Evidence-Based Policymaking Act of 2018, requires federal agencies to make data accessible, to develop statistical evidence to support policymaking, and report annually to the Office of Management and Budget and Congress. There are many private sector efforts, e.g., [Evidence based Policymaking](#), [Coleridge Initiative](#) with analysts recommending that organizations use data-driven techniques to re-engineer all decision making[38].

The greatest challenges of using data science to improve human decision making is the triple threat inherent in data science – inscrutability and challenges of using data science to formulate and solve the decision making problem. Consider ensuring that the patterns discovered in a dataset are factual, i.e., occur in sufficient frequency to be included in the data science result. The value of such data-driven results depends on many factors. The trained, tuned data science model must conduct the intended analysis. The dataset must adequately represent both the phenomenon's features and the population that are critical to the analysis. The explanation of the solution and the interpretation of the results must be verified and validated to the degree required by the motivating domain problem. Finally, the analysis must be data-driven, requiring careful acquisition and preparation of the dataset to attempt to ensure that it correctly represents the features of the phenomena that are critical to the analysis. The raw, observational data must be unchanged except for error correction and representation to improve data engineering without impacting information content. Due to the challenges of designing, developing, and tuning data science solutions, and to the no *free lunch conjecture* (§2.13, [44]), improving confidence that the results reflect reality requires experimentation with data science solutions and datasets. Understanding the probabilistic nature of data science results should guide the interpretation of data-driven results, recalling that data science results are not answers to a domain problem but insights into the nature of the domain problem. Finally, data-driven decision making is *post-facto*. To determine the potential results of a decision, e.g., government policy, the decision had to have been in place for the evidence of its effect to be discovered in a dataset. Lacking the above due diligence can pose significant challenges in applying data science.

### *2.12. Beyond human reasoning*

Data science operates outside the realm of human, mathematical, and scientific reasoning, as it is devoid of human concepts, thoughts, and reasoning, as far as we know. We do not know how or what data science models learn or infer from datasets. We use human terms – learn, infer, explain, interpret, discover, reason – to describe data science as rough analogies that may be misleading or incorrect, especially with respect to the scope, scale, and complexity of data science problem solving. For example, data science image recognition using convolutional neural networks work perfectly well on images that the human eye could not discern.

The lack of human concepts and reasoning may be one of data science's greatest benefits. Data science is not limited by human scope, scale, and complexity, nor by the Cartesian partitioning of human reasoning. Hence, data science may also be devoid of such human limitations. For example, *bounded rationality* applies in disciplines such as political science, economics, psychology, law, and cognitive science in which human rationality is limited by the difficulty of the problem, cognitive ability, lack of information, time to decide, etc. causing rational individuals to select a decision that is satisfactory rather than optimal. Similarly, *prospect theory,* by Daniel Kahneman (2002 Nobel Prize in Economics) and Amos Tversky, has been proven in practice and using ML to explain limitations of human decision making in the face of economic risk and too many competing probabilities for the brain to accommodate. Being devoid of human reasoning, data science may be devoid of such limitations. Data science provides fundamentally different means for *understanding* and *analyzing* phenomena. Hence, data science can discover patterns that human concepts and reasoning cannot. Data science can be used to evaluate scientifically established knowledge for alternative insights and solutions. An advantage of data science's non-human nature is that to discover patterns relating to human concepts, e.g., lobsters in images, words in language, or values of properties such as uncertainty, bias, fairness, and equity, domain problems must be translated into data science problems in a form that data science can discover. This may require seeing well-known human concepts and values more clearly in terms of their essential properties, e.g., what patterns characterize equitable healthcare? That is, data science may require understanding the world in new, unconventional ways, as is widely recognized. "And as AI becomes more widely used and makes findings that surpass human understanding —whether concerning the laws of science, medicine, managing businesses or navigating roads— society may seem at once to be hurtling towards knowledge and retreating from it."[19]

The greatest challenge of effectively using data science being devoid of human concepts and reasoning is inscrutability. We do not understand how it works. As there is no theory of data science, discovering how best to use it involves considerable exploration of the problem and solution spaces. Understanding phenomena outside human reasoning is difficult. For example, for a century, physicists have attempted to unify Einstein's general and special relativity. Clearly the world is unified, but humans have not been able to reason across these human-created concepts and theories. How might data science be used to discover a unified view of the world in sharp contrast to our artificial, human Cartesian view? The common claim that data science is multidisciplinary is false. Data science is devoid of human defined disciplines. Multidisciplinary is a human approach to better understand and use data science. It is challenging to acquire and prepare a data science dataset that truly represents those features of a phenomenon essential to an analysis when those features result from human versus a non-human, unified perception of the phenomenon.

"Living with this technology will be tough. Based, as they are, on correlations and elaborate statistics, rather than on a sense of causality, AI's decisions may seem otherworldly; when the stakes are high, they must be diligently validated." [19]

### 2.13. Data science solutions

A data science solution with results that have been demonstrated to have credible explanations and interpretations for a motivating domain problem can be applied to any data science dataset that meets the requirements for the analysis. In contrast, scientific knowledge resulting from verified and validated scientific analyses are considered universal by scientific induction, hence are repeated only to demonstrate repeatability. Successive applications of a data science solution are often modified, improved with additional training and tuning. Every dataset unseen by a data science solution produces new results. There is no final or singular result for a data science problem due to its probabilistic nature. A chief benefit of a data science solution is reusability – over time, libraries of established solutions can be reused to gain new insights into motivating domain problems using different datasets. Such solutions, e.g., foundation models[2], are seen as the future of applied data science. A remarkable benefit of foundation models is that they can be used to conduct tasks, called emergent properties of a model, for which they were not intentionally trained.

### 2.14. Data science principles and laws

Some of the most valuable scientific discoveries are expressed as scientific principles that if they are quantifiable, as scientific laws, that approximate knowledge of the natural world. For example, $E=MC^2$ and $F=MA$ are scientific laws that approximate measures of energy and force in physics. Such laws encapsulate scientific knowledge and have provided understanding for centuries. For similar reasons, It would be useful to have data science principles and laws that capture essential, but probabilistic, knowledge of phenomena. Consider candidate data science principles and laws.

*For information to be discovered, it must be in the dataset* is data science law that is inherently impossible to achieve or prove.

*The properties of a data science model are determined by the properties of the computational method, modified by training and tuning techniques, and by the dataset and its preparation techniques. Ideal model and dataset preparation cannot modify the information (knowledge) inherent in the dataset and may not be able to discover it. The ideal is seldom achieved and can be assessed only outside data science, e.g., empirically.*

*No free lunch conjecture:* Based in part on David Hume's 1739 reasoning that cast doubt on inductive reasoning[14] – a tenant of science, some mathematicians believe that for a given analysis and analytical dataset, there is no optimal algorithm (function, accuracy, performance, resource consumption, etc.). There is no optimal choice amongst hundreds of methods and thousands of algorithms[4] for analyzing a dataset.

---

[4] BlobCity AI Cloud provides 1,000+ ML template algorithms for classification, regression, clustering, EDA, dimensionality reduction, time series analysis, NLP, and audio-visual methods.

This claim, known as the *no free lunch theorem*, states that there are no optimal computational method or algorithm for a specific analysis of a specific dataset. It has been proven in many cases, e.g., Wolpert and Macready[44] under qualified conditions. Arguments against *no free lunch* are based on proofs of unrealistic cases, e.g., clean, pristine datasets. No free lunch suggests the extreme difficulty, if not impossibility, of finding an optimal method or algorithm. Based on a career with data and observational data science datasets, I intuitively extend the no free lunch theorem to the *no free lunch conjecture*: *First, there is no single algorithm that produces an optimal result for a specific data science model and dataset. Second, there is no optimal, observational dataset for a data science analysis of a phenomenon*. These conjectures cannot be proven but provide guidance. Any truly observational dataset is unlikely to truly represent the phenomenon, no matter how large the dataset. For example, despite the world-wide focus and advanced technology used for COVID-19, there is no dataset with adequate data to explain why blacks were 1.3 times and Hispanics 2.4 times more likely to be infected as white Americans[6]. However, despite the cause, those insights led to more effective vaccines. Acquiring the ideal dataset for a specific data analysis, i.e., with the highest entropy, is statistically unlikely to impossible[5]. This applies to every selection for an algorithm and dataset for every data analysis. Having developed a data science solution with validated explanations and verified interpretations of the motivating domain problem, when do you stop training and tuning the data science model and acquiring and preparing a data science dataset? You don't. Data science is not capable of certain, proven outcomes. It merely provides insights.

*Data science is no more about data or analytical methods than astronomy is about telescopes[6].* Data and methods are means for discovering what Tukey called *lies beneath the data*. Data science discovery is about uncertainty, not human-preferred certainty.

Data science is not even about the phenomenon being analyzed. *Data science is about discovering plausible, fact-based but probabilistic evidence of patterns of features of a phenomenon by analyzing observational data intended to represent the phenomenon* – features that the analyst may never have considered.

*The inverse power law - exponential resource consumption for linear gain:* The cost to improve large language models has grown exponentially with linear improvements in the models, e.g., BERT (16 GB of data, 3.3B words 345M parameter, $7K cost) beaten by GPT-3 (45TB of data, 0.5T words, 175B parameters, $4.6m), continuously beaten by larger, more costly models. Empirical evidence suggests that the performance of language models increases as a power-law with the number of parameters (model size), dataset size, and computational budget. OpenAI suggests that the amount of compute used for training AI models doubles every 3.4 months with negative consequences for the environment and economy, i.e., a few commercial organizations control the best models. This tide seems to be turning.

### 2.15. Data science datasets – greatest strength and weakness

Data science datasets are a great strength of data science, when they contain patterns that provide humans insights into the domain problem, solution, and result, and a great weakness when they do not. There is no free lunch to get the perfect, unattainable dataset or at least there is no way of determining if it is best, i.e., has the highest entropy relative to the intended analysis. The inverse power law suggests that larger dataset are better at higher cost. Good datasets are hard to acquire and impossible to evaluate other than empirically. The more control placed on acquiring and preparing the dataset, the worse the result –

---

[5] Invariably, for every obtained dataset *the entropy horse has already left the discovery barn*.

[6] After Dijkstra's famous quote.

the less likely is the value of the result, and the less the analysis leads to valuable insights.

### 2.16. The multiple definitions challenge

As it was with science in its infancy, data science is emerging rapidly and simultaneously in 40+ disciplines, hundreds of disciplines, and tens of thousands of applications resulting in millions of publications. These often independent developments result is multiple definitions and terminology for similar or identical concepts. The resulting *multiple definitions challenge is* complex and poses challenges that can be addressed by developing a unified understanding and definition of data science. The problem, challenges, solutions and benefits proposed in[46] are summarized below.

The chief drawback of the *multiple definitions challenge* is that while data science is understood and applied adequately in most data science disciplines, it is inadequately understood and defined as a field of inquiry with analytical power greater than science. This limits understanding, communication, and collaboration across disciplines employing data science thus limiting the development of data science as a field of inquiry and as a general purpose technology[12]. A potential challenge to the development of data science is Kuhn's *incommensurability thesis[22][39]* that concerns multiple definitions of science but applies to all research paradigms including data science. Multiple definitions are integral to an emerging, evolving field of inquiry reflecting variations across disciplines and advances as it matures. The incommensurability thesis suggests that data science, data science disciplines, and data science results obtained under different definitions of data science may be incomparable without means of translating between them. In the worst case, one data science definition may invalidate not only another definition, but also the results achieved with that definition. This may seem extreme, however, the more complex challenge of determining the coherence of two definitions of data science is real. This research [46]-[54] provides means for developing and maintaining coherent definitions of data science, data science problem solving, and data science disciplines, hence, incommensurability.

### 2.17. Unfathomed utility

Due to data science's many benefits, especially the unfathomed scope (applicability, a general purpose knowledge discovery paradigm), scale, complexity, and power, it has been successfully applied to solve strategic and practical problems in applications in most disciplines, in most industries, and in new data science-based industries and industrial sectors such as bioinformatics. After Two Sigma's 2001 adoption of data science-based trading in the quantitative investing industry, such strategies emerged rapidly. Early enormous wins (44%) were followed, in 2019 by underperformance. By 2020, data science had revolutionized trading as means to discover opportunities for human traders. By 2023, it dominated the investment industry[24] with humans in the loop due to the inherent risks.

The importance and value of the realized successes has led to extensive visibility and widespread adoption and demand for data science based products, services, and skilled employees. There are extensive government, industry, and academic reports on the value and importance of data science applications[37] and on consequent threats, e.g., to national security[26][27]. This section cites assessments by world experts and presents a summary of the nature and benefits of data science adoption and deployment. Strategic and practical achievements made while data science is in its infancy suggest its potential for unfathomed value.

An indication of the importance of AI-based data science is seen since 2020 in it being called Artificial Intelligence (AI) as if data science had eclipsed all AI sub-fields including evolutionary computation, vision (image recognition), robotics, expert systems, speech (natural language) processing, and planning. This is true of all cited references. In early 2023, AI referred even more narrowly to generative AI.

In 2022-2023, most popular publications, e.g., Time, Fortune, declared the importance and value of AI-based data science, e.g., The AI Era begins right now[21]. "Bill Gates says what's been happening in AI in the last 12 months is *every bit as important as the PC or the internet*" [Time, January 2023]. The importance and value of the AI-based data science has been compared with the steam engine and electricity and as important as the Fourth Industrial Revolution [Forbes Nov 7, 2019]. In 2017, Sundar Pichai, CEO of Google's parent company Alphabet, said "AI was one of the most important things humanity was working on, … more profound than … electricity or fire."[4] "A new industrial revolution has begun. Like mechanization or electricity before it, artificial intelligence will touch every aspect of our lives—and cause profound disruptions in the balance of global power, especially among the AI superpowers: China, the United States, and Europe"[34], now called the AI Arms Race.

The importance and value of AI-based data science warranted studies by most governments and international bodies, including the World Economic Forum (§5). The US government conducted many such studies in many areas including national security[26] that observed "AI technologies will drive waves of advancement in critical infrastructure, commerce, transportation, health, education, financial markets, food production, and environmental sustainability" and that AI is an engine of innovation with examples in every sector[26]. Studies[26][27] call for a national strategy for *Intelligence* as a new industrial macroeconomic sector like transportation, energy, manufacturing, finance, and economics. They advise governments and industry to make large-scale investments in this emerging sector, as Europe did in physics (CERN), to improve products and services, to stimulate innovation, and for America to maintain its lead in data science especially over China that has made the largest investments of any country after the USA. In 2022, China surpassed the USA in refereed data science publications and often matches major US data science achievements.

There is an intense worldwide competition to develop AI and translate its results into not just economic, cultural, and military power, but to establish a virtuous cycle of continued innovation.

Most nations are executing on established AI strategies (§6). "In 2017, China laid out plans to become the world leader in artificial intelligence (AI) by 2030, with the aim of making the industry worth 1T yuan ($147.7B)"[3]. In February 2019, the White House issued Executive Order 13859, *Maintaining American Leadership in Artificial Intelligence*. In 2020, the US Congress passed the National Artificial Intelligence Initiative Act of 2020 (NAIIA) (DIVISION E, SEC. 5001)[37] (§7). In 2023, Eric Schmidt advised the US government to invest heavily in AI to achieve *Innovation Power* – the ability to invent, adopt, and adapt new technologies, that will restructure the geopolitical landscape[35]. To demonstrate AI's Innovation Power, he cited Ukraine's use of AI to successfully resist the military giant, Russia[35], in this case in the service of democracy. While the innovation power of technology has led to success in the past, the "sheer speed at which innovation is happening has no precedent. Nowhere is this change clearer than in one of the foundational technologies of our time: artificial intelligence" [35].

The potential value and importance of data science is a topic widely and frequently published by experts in relevant domains. The following are examples of the axiology (practical and strategic value) of data science.

**Startups and investing**: Belief in the potential of data science can be seen in investments. In early 2023, NFX Ventures, a VC firm, lists 539 generative-AI startups with over $11B invested on top of $11B invested by Microsoft in OpenAI. Generative-AI is only one of 20 data science disciplines or technologies[40]. Its high visibility is due to products and services based on OpenAI's ChatGPT that has been matched by similar products and services from most leading US and Chinese AI organizations. Generative-AI is rapidly being deployed in the arts and entertainment industry and is projected to transform and disrupt web search where it is deployed with billions of users daily.

**Demand and adoption:** In the late 2010s, most large US organizations declared data science to be a core strategy, deployed it to take advantage of vast amounts of information for strategic planning, then to improve operations, and now to produce products and services. In 2017, Google declared its cloud strategy to be "AI first"[25]. Data science often exceeds human performance, e.g., medical imaging, leading to incorporating data science solutions into operations. ChatGPT set the record for the fastest technology adoption in history reaching 100 million active users in two months with 16 million daily users. While ChatGPT has captured the world's attention, it is an inaccurate indicator of the data science axiology.

**New products, services, and sectors**: Data science has led to Intelligence as a new macroeconomic sector[26][27]. New data science disciplines led to new sectors within existing industries, e.g., fintech, bioinformatics (digital, computational biology), digital history and archeology (e.g., cliodynamics), digital social science[23], medical therapeutics[13], digital medical phenotyping that extracts real-world evidence from electronic medical records and doctors' notes to accelerate medical analysis and treatment planning. Data science is used to address existential challenges, notably climate change[33]. Generative-AI is transforming web search, coding (80% of new code by products like OpenAI's Codex, GitHub's Co-Pilot, Google's AlphaCode, Salesforce's CodeGen), writing essays, and entertainment (DALL–E 2). This is just the beginning.

**Applied data science**: Foundation models are "any model that is trained on broad data (generally using self-supervision at scale) that can be adapted (e.g., fine-tuned) to a wide range of downstream tasks"[2]. They are data science models that have demonstrable, standard data science solutions, i.e., demonstrated to meet properties like accuracy, reliability, and robustness for specific domain problems. Initially, foundation models were developed for language (NLP), vision (image recognition), robotics, reasoning, and human interaction with many more application classes to follow. The principle is that the behavior of the underlying data science models are adequately understood, operate the same way in many domains (homogenization), with known strengths and weaknesses, e.g., single points of failure. Foundation models can be adapted to a wide range of downstream tasks hence are a "paradigm for building AI systems". They are seen as the principal means for developing and deploying safe, trustworthy data science applications.

The chief benefit of data science's unfathomed utility is that its proven and potential use to solve problems for any phenomenon for which an adequate data science model and data science dataset can be created. As a general-purpose technology, like electricity and information technology, data science has the potential to drive innovation, competitiveness and productivity in every industrial, academic, and government sector[26][27]. Chief among the many proven and potential benefits described here, is its ability to address problems of a scope, scale, complexity, and power greater than any other problem solving paradigm. Consulting firm PwC projects that AI will increase global GDP by up to 14 percent ($17.5T) by 2030 due to its ability to drive productivity gains by automating business processes and augmenting human labor. Consulting firm Accenture estimates that AI could increase US labor productivity by 35 percent and increase the annual growth rate of gross value added to the U.S. economy from 2.6 percent to 4.6 percent by 2035 through automation and improving labor and capital management.

### 2.18. *Unfathomed risks - continued*

Data science's unfathomed utility is juxtaposed with its unfathomed risks that warranted US government national security studies to identify and assess them. Inherent in data science is the potential of leading to erroneous results in addressing motivating domain problems. Examples include racially insensitive remarks (Microsoft Tay, 2016), plagiarism, and misinformation, e.g., CNET's use of a proprietary generative-AI model to produce personal financial advice, and MetaAI's Galactica that produced erroneous scientific content, racists comments, and convincing nonsense. Perhaps worse is data science's power to harm (§2.6). Highly visible examples are misinformation and deepfakes – a deep learning application in which an image, video or audio of a person is replaced with someone else's likeness.

The utility of data science raises many challenges. Data science's power to increase productivity and outperform humans in many tasks leads to concerns that data science may take jobs from humans. Despite epochal, technological, and economic change, fears of mass technology-based unemployment have never been realized not even in the face of the current AI boom[35]. Experts suggest that the USA has too little versus too much automation[7]. Data science's potential may change those results. Laws and regulations governing new technologies to maximize benefits and limit harms lag technology advances often by decades as the nature and impact are understood. Three such issues have arisen for ChatGPT. First, generative-AI platforms may not enjoy the legal protection from liability that shields social media. Second, copyright holders of web-based content, e.g., Getty images, on which existing models have been trained without permission or compensation have filed lawsuits against generative-AI product companies for unauthorized usage. Third, generative-AI products may not ensure legal and privacy protection on data that they acquire and generate. Fourth, generative-AI products may not enjoy copyright protection that currently pertains to human generated content. Finally, there is concern that profits and growth may take precedence over safety in deploying data science products and services. These concerns have led research and practice of ethical and trustworthy AI that is lawful, ethically adherent, and technically robust.

**3. Data science axiology**

This section summarizes the axiology of data science that can be seen in contrast with science. The scope, scale, and complexity of problems addressed by data science and its power to gain insights into those problems vastly exceed those of science, as described in §1.2 and [46]. First, science and data science are general purpose research paradigms that can be used within other research paradigms, and *vice versa*. Second, while science has contributed to reasoning and problem solving in certainty where humans are most comfortable, data science contributes to reasoning and problem solving in uncertainty that may be more realistic. Third, scientific analysis relies exclusively on human concepts and reasoning with its attendant limitations while data science is devoid of human concepts and reasoning and their limitations, e.g., developing a unified field theory in physics. Fourth, while scientific problem solving is challenging, data science is more challenging since in addition to formulating and solving a data science problem, the results must be used to solve the motivating domain problem by the yet to emerge wealth of such solution techniques. Fifth, by scientific induction, scientific results are universal requiring no repetition other than to demonstrate repeatability while data science solutions and their results are probabilistic and are intended for continuous problem solving with unseen datasets representing different phenomena possibility using data science models improved by additional training and refinement. Sixth, important, proven scientific knowledge is expressed in elegant scientific principles, or if quantifiable, in scientific laws (e.g., $E=mc^2$, $F=MA$) that approximate reality, i.e., ignore or cannot detect some details, substantiated in 1927 by Heisenberg's uncertainty principle. Accepting scientific theories as *inherently true* of nature, as is common, overlooks the limitations of scientific reasoning and theories. In contrast, data science can discover, probabilistically, all significant patterns, causing some to declare that data science is in a *post-theory world – figure it out on demand*! Imagine seeing the world in terms of all relevant facts rather than via approximate theories! Which is more real? more useful? Seventh, since science exclusively considers deterministic outcomes, it cannot identify valuable, more realistic, probabilistic outcomes discoverable with data science. For example, during the COVID-19 pandemic, Moderna, a leading COVID-19 vaccine producer, used data science to analyze data obtained in randomized control trails to discover probabilistic COVID-19 prognoses that led to vaccines for missed cohorts thus accelerating that prognosis and saving millions of lives.

An eight and final difference requires examples. Scientific experiments allow no degrees of freedom while data science enables many. To provide definitive results, scientific experiments rigorously follow a validated empirical design to analyze a small, bounded number of variables for a single class of well-defined phenomena. In contrast, data science is used to conduct analyses over an unbounded, typically unknown, number of variables in datasets that can represent similar but not identical phenomena. For example, international investing requires analyzing and comparing globally traded financial instruments in different jurisdictions that must satisfy different criteria in each jurisdiction. Similarly in quantitative economics, Piketty's Capital in the 21st Century[30][31] analyzed income from assets versus income from labor over observational data collected from the economies of over 100 countries over 100 years. The dataset was observational since it was collected under controls that varied over time across different economic models, reporting, and metrics across and within countries. The analysis concluded that income from assets grows faster than income from labor[7] with an accuracy of 0.1 percent. What is critical here is that ultimate truth was determined not by computation but by consensus of the economics community. This type of analysis is common for estimating financial risk in the banking and investments communities for instruments that are denominated and governed in different regulatory regions making apples-to-apples comparisons impossible. Previously done by humans, it is now done with data science to improve risk estimates and to accelerate analysis of the heterogenous phenomena. Another example of degrees of freedom in data science that does not exist in science is the use of ensemble methods to combine analytical results from multiple data science models to obtain analytical results that surpass those of a single model.

### 4. Data science is changing our world

The greatest benefit of data science is its proven ability to address problems of scope, scale, and complexity beyond human capacity to understand and not otherwise possible, combined with its analytical power and speed. In these and other properties, it exceeds all research paradigms. Data science offers hope of addressing man's greatest challenges previously not possible. The benefits depend on sound, trustworthy, ethical use of data science. The greatest challenge of data science is its inscrutability leading to many challenges most significantly its unfathomed risks of omission and commission. Ideally, current practical technical solutions will be augmented, ideally with provable theories. Data science will be seen as a milestone in the development of reasoning that will profoundly change human society, as seen by world leaders in technology, history, and statesmanship, "*the most important way that AI will change society is by redefining the basis of knowledge*" and "*What does it mean to be human?*" [10][11][19][20][35].

Technology has led to epochal impacts on and changes to society and life. 15th century printing presses led western society from the Age of Religion, in which the church dominated knowledge and knowledge discovery, to the Age of Reason – The Enlightenment – in which knowledge and knowledge discovery became the providence of educated people through rigorous scientific, political, and philosophical human-centered reasoning. 400 years ago the scientific revolution profoundly changed knowledge discovery and led to fundamental changes in society and the world order. At that time, automation launched the industrial revolution that transformed society without impacting knowledge discovery. The Information Age improved knowledge management 200 years later without impacting knowledge discovery until the emergence of AI-based data science at the turn of the 21st century. Over the coming decades, data science will be recognized as the most powerful research paradigm in history exceeding the power and axiology of science. The Age of Enlightenment fundamentally changed human society and human life. How will the Age of AI change our world?

---

[7]The first proof in economics of the age-old aphorism that the *rich-get-richer*.

Seen from science, the Enlightenment held that the universe was finite, certain, and understandable in human concepts. Scientific evidence then proved that view to be wrong. The universe is unfathomed and uncertain probably to the chagrin of many. The Age of AI, devoid of human concepts and reasoning, may take human society into a post-theory world understood in terms of probabilistic patterns with significance, determined by data science that is currently inscrutable but critical to study. Humans have no philosophical basis for understanding data science. Is its cognition fundamentally different from human cognition? Is there a metacognition that we have yet to conceive? Data science is changing our understanding of reality. "Whether we consider it a tool, a partner, or a rival, AI will alter our experience as reasoning beings and permanently change our relationship with reality" [19]. In science, truth is provable. In philosophy, truth is rationally related to facts – both defined by humans. Data science has no human concept of truth. Many speculate that the world is unprepared for such change. How will The Enlightenment end? Is human society prepared for the rise of AI[18]?

**5.     Epilogue: Our Promethean AI moment**

AI-based data science can be used to discover knowledge and solutions to problems beyond human capacity in scope, scale, complexity, power, and speed that could not be gained otherwise. Recently, its unfathomed power has been doubling every few months, leading to a Promethean moment that will profoundly change our world. As in Greek mythology when Prometheus defied the gods by giving mankind fire – means to create and destroy – we face the challenge of using AI-based data science for good or for evil. There are typically two solutions to ensure positive, safe, ethical uses of technology – by humans and by technology. Due to inscrutability, there are no technical solutions. None may arise. Meanwhile, this inscrutable technology is being adopted and applied faster than any technology in history.

Our Promethean challenge is to maximize positive benefits while preventing harmful, destructive uses. Science, the previously most powerful knowledge discovery paradigm, took 400 years to formalize. Data science, a far more powerful and dangerous paradigm, may take decades, but poses imminent, urgent risks. The Promethean challenges of the Age of Enlightenment were addressed by the greatest minds of the age. The Promethean challenges in the Age of AI should be considered by the great minds of our age.

*Open the pod bay doors, HAL.*
*I'm sorry, Dave. I'm afraid I can't do that.*
*2001:* A Space Odyssey (1968)

**6.     Appendix I: World Economic Forum on AI**

The World Economic Forum is an independent international organization committed to improving the state of the world by engaging the foremost political, business, cultural and other leaders of society to shape global, regional and industry agendas. It explores and monitors the issues and forces driving transformational change across economies, industries and global issues including artificial intelligence. The World Economic Forum© figure (accessed March 10, 2023) depicts Artificial Intelligence, a new industrial sector, as a basis for an axiology of AI in industries, economies, and values.

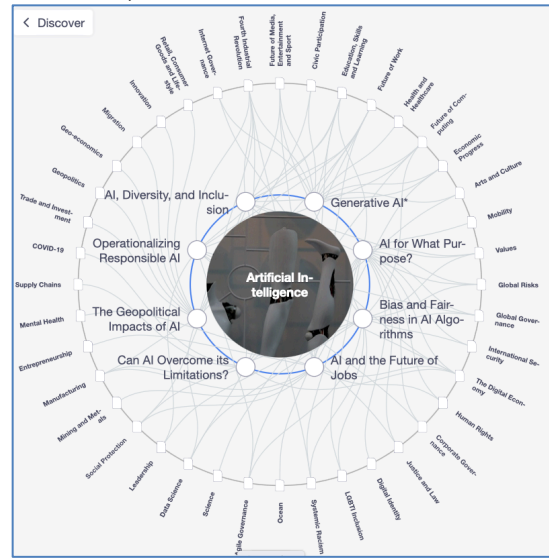

**7.     Appendix II: International AI strategies**

Starting in October 2021, Brookings Institute identified 44 countries that announced AI strategic plans and monitors and ranks their fulfillment of those plans. See Fig 1 Four quadrants of national AI implementation.

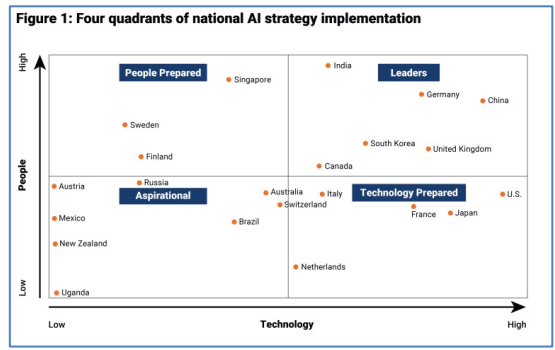

## 8. Appendix III: US National Artificial Intelligence Initiative Act of 2020

From [National Artificial Intelligence Initiative](#) website (accessed March 18, 2023).

The National Artificial Intelligence Initiative (NAII) was established by the National Artificial Intelligence Initiative Act of 2020 (NAIIA) (DIVISION E, SEC. 5001) – bipartisan legislation enacted on January 1, 2021. The main purposes of the initiative are to ensure continued US leadership in AI R&D; lead the world in the development and use of trustworthy AI systems in public and private sectors; prepare the present and future US workforce for the integration of artificial intelligence systems across all sectors of the economy and society; and coordinate ongoing AI activities across all Federal agencies, to ensure that each informs the work of the others.

The National AI Initiative Act of 2020 (DIVISION E, SEC. 5001) became law on January 1, 2021, providing for a coordinated program across the entire Federal government to accelerate AI research and application for the Nation's economic prosperity and national security. The mission of the National AI Initiative is to ensure continued U.S. leadership in AI research and development, lead the world in the development and use of trustworthy AI in the public and private sectors, and prepare the present and future U.S. workforce for the integration of AI systems across all sectors of the economy and society.

In support of the Initiative, the NAIIA directs the President, acting through the NAIIA Office, interagency committee (Select Committee on AI) and agency heads, to sustain consistent support for AI R&D, support AI education and workforce training programs, support interdisciplinary AI research and education programs, plan and coordinate Federal interagency AI activities, conduct outreach to diverse stakeholders, leverage existing Federal investments to advance Initiative objectives, support a network of interdisciplinary AI research institutes; and support opportunities for international cooperation with strategic allies on R&D, assessment, and resources for trustworthy AI systems.

The National AI Initiative provides an overarching framework to strengthen and coordinate AI research, development, demonstration, and education activities across all U.S. Departments and Agencies, in cooperation with academia, industry, non-profits, and civil society organizations. The work under this Initiative is organized into six strategic pillars – Innovation, Advancing Trustworthy AI, Education and Training, Infrastructure, Applications, and International Cooperation.

**Acknowledgement**
I thank Prof. John Mylopoulos, University of Toronto, for thoughtful comments on this paper.

**Data science reference framework**